\documentclass[letterpaper, 10 pt, conference]{ieeeconf}  % Comment this line out if you need a4paper

\usepackage{comment}
\usepackage{graphicx}
\usepackage{hyperref}
\usepackage{cite}
\usepackage{amssymb}
\usepackage{amsmath}
\usepackage{mathtools}
\usepackage{mathastext}
\usepackage{balance}
\DeclareMathOperator*{\argmin}{arg\,min}

\usepackage[linesnumbered,ruled,noend,vlined]{algorithm2e}
\SetArgSty{textup}
\usepackage{xcolor}

\SetCommentSty{mycommfont}
\bstctlcite{IEEEexample:BSTcontrol}

\IEEEoverridecommandlockouts                              % This command is only needed if 
\title{\LARGE \bf \textit{CollisionIK}: A Per-Instant Pose Optimization Method for Generating Robot Motions with Environment Collision Avoidance}

\author{Daniel Rakita, Haochen Shi, Bilge Mutlu, Michael Gleicher
\\
\normalsize{Department of Computer Sciences, University of Wisconsin--Madison}\\
\{rakita,hshi7,bilge,gleicher\}@cs.wisc.edu
}
\begin{document}

\maketitle
\thispagestyle{empty}
\pagestyle{empty}

%%%%%%%%%%%%%%%%%%%%%%%%%%%%%%%%%%%%%%%%%%%%%%%%%%%%%%%%%%%%%%%%%%%%%%%%%%%%%%%%

%%%%%%%%%%%%%%%%%%%%%%%%%%%%%%%%%%%%%%%%%%%%%%%%%%%%%%%%%%%%%%%%%%%%%%%%%%%%%%%%

\begin{abstract}
In this work, we present a per-instant pose optimization method that can generate configurations that achieve specified pose or motion objectives as best as possible over a sequence of solutions, while also simultaneously avoiding collisions with static or dynamic obstacles in the environment.  We cast our method as a multi-objective, non-linear constrained optimization-based IK problem where each term in the objective function encodes a particular pose objective.  We demonstrate how to effectively incorporate environment collision avoidance as a single term in this multi-objective, optimization-based IK structure, and provide solutions for how to spatially represent and organize external environments such that data can be efficiently passed to a real-time, performance-critical optimization loop.  We demonstrate the effectiveness of our method by comparing it to various state-of-the-art methods in a testbed of simulation experiments and discuss the implications of our work based on our results.
\end{abstract}
\section{Introduction}
\label{sec:intro}

Optimizing individual poses to achieve some objective is a common technique in robotics.  For example, inverse kinematics (IK) is one instance of single pose optimization that involves optimizing a robot's joint-angle pose to match an input 6-DOF end-effector position and orientation goal.  Single pose optimization has many practical benefits, such as its speed and comprehensible ``single-input-to-single-output'' nature.  However, this paradigm is often insufficient when used iteratively to generate motions over a sequence of poses, a paradigm historically referred to as \textit{per-frame IK} in animation \cite{lee1999hierarchical, gleicher2001comparing}.  Problems with per-frame (or, per-instant) optimization often stem from the stream of independent poses lacking \textit{temporal coherence}, \textit{e.g.}, jerky motion or joint space discontinuities, or lacking \textit{motion feasibility}, \textit{e.g.}, self-collisions or collisions with obstacles in the environment.  Thus, despite the speed and convenience of per-instant pose optimization, its shortcomings mean that larger motion planning or trajectory optimization frameworks are often needed in order to achieve coherent and feasible motions.                   

\begin{figure}[t!]
	\includegraphics[width=\columnwidth]{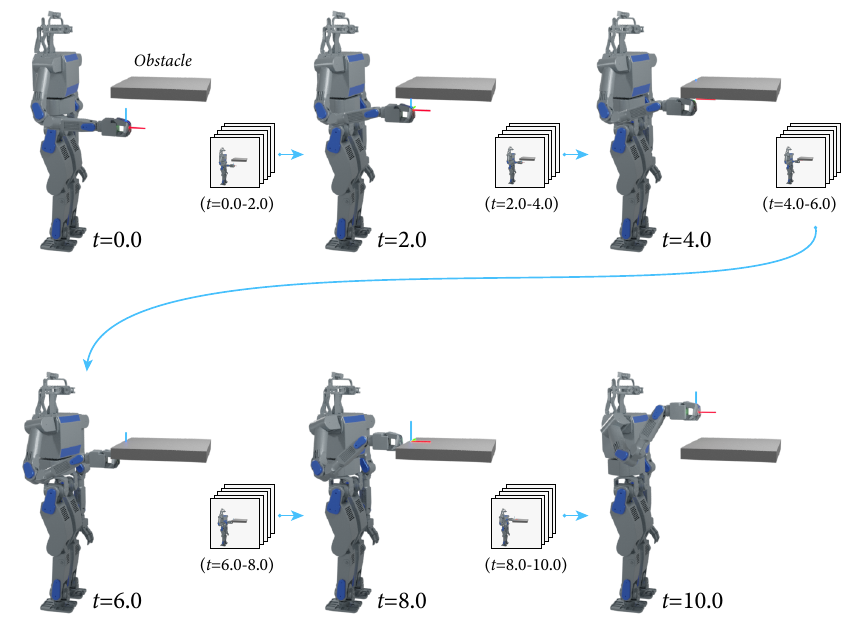}
	\caption{In this work, we present a per-instant pose optimization method that is able to generate smooth, feasible paths on-the-fly without the use of a planner or trajectory optimizer.  This example shows a robot avoiding a collision with a table while maintaining the same end-effector orientation throughout the motion.}
	\label{fig:teaser}
	\vspace{3pt}
\end{figure}

In this work, we present a per-instant pose optimization method, called \textit{CollisionIK}, that optimizes single poses at given time points that achieve certain accuracy objectives as best as possible, \textit{e.g.,} matching end-effector pose goals, without sacrificing temporal coherence and motion feasibility on a sequence of solutions.  In particular, our method features environment collision avoidance as a sub-component, meaning that feasible motions that avoid collisions with static or dynamic obstacles can be generated on-the-fly in a per-instant manner without requiring a motion planner or trajectory optimizer.  Our method can also incorporate joint smoothness objectives that use a short history of previous solutions to optimize over approximated derivatives.  This per-instant paradigm, where smooth motions are iteratively constructed on-the-fly, effectively allows the robot to react and adapt its motion to complex or dynamic environments, all while still reflecting other objectives as best as possible.     

We cast our method as a multi-objective, non-linear constrained optimization-based IK problem.  The objective function is a weighted sum, where each term in the sum encodes a particular motion objective.  The weights on the terms in the objective function set importances of the various terms and allows the optimization solver to \textit{relax} certain objectives in favor of other, more important, terms in the case of competing objectives.  In particular, our method favors motion feasibility objectives, such as avoiding collisions, over other objectives, such as matching end-effector pose goals.  Throughout this work, we overview the structure of this overall optimization framework and highlight how this or similar frameworks extend well to the case of environment collision avoidance. 

% In the context of robot motion generation, this approach builds on the paradigm called \textit{relaxed} inverse kinematics (RelaxedIK) \cite{rakita2018rss}

Our current work offers two technical contributions: (1) we demonstrate how to effectively incorporate environment collision avoidance as a single term in a multi-objective, optimization-based IK structure (\S\ref{sec:objective_term}); (2) we provide initial solutions for how to spatially represent and organize external environments such that data can be efficiently passed to a real-time, performance critical optimization loop (\S\ref{sec:env_representation}, \S\ref{sec:collision_scene_filtering}); and (3) we provide open-source code that implements our proposed method: [\textit{link will appear here}].    

We assessed the efficacy of our method by running a testbed of simulation experiments (\S\ref{sec:evaluation}).  We compared our method to the MoveIt! library Cartesian path controller \cite{chitta2012moveit} and \textit{RelaxedIK} \cite{rakita2018rss} on various simulated robots and tasks.  We demonstrate that our method successfully avoids collisions with static or dynamic objects in the environment in real-time while consistently achieving additional motion objectives.  Our evaluation also shows that our method scales well and maintains its efficient performance even in environments with many obstacles.  We discuss the benefits and drawbacks of our method compared to the alternative approaches, such as speed, local minima, and motion accuracy trade-offs, and conclude with an overall discussion about the implications of our work based on our results (\S\ref{sec:discussion}).

\section{Related Works}
\label{sec:related_works}
In this section, we highlight prior works that our method draws inspiration from in the areas of motion-planning, trajectory optimization, animation, and inverse kinematics.

\textit{Motion Planning}---
Generating robot motions that avoid collisions with obstacles in an environment is commonly addressed with a technique called \textit{motion planning} \cite{lavalle2006planning}.  This approach finds collision-free, feasible paths from a start state to a goal state in configuration space.  Sampling-based motion planners often use random samples to bootstrap a search strategy and build a graph structure from start to goal \cite{lavalle1998rapidly, kavraki1994probabilistic, hsu1997path, kuffner2000rrt}.  Such planners are commonly guaranteed to find a solution if one exists, and some variants, such as RRT* \cite{karaman2011sampling}, are also guaranteed to find the \textit{shortest} feasible path in the limit. 

While standard planners are effective at eventually finding collision-free solution paths, provided one exists, they are less adept at consistently finding sufficient solutions in a time-sensitive, real-time setting.  Variants of these approaches have focused on the real-time aspects of this problem, such as Kroger et al. \cite{kroger2009online}, who presented a real-time planning algorithm that allowed robots to avoid dynamic obstacles in real-time.  Hauser \cite{hauser2012responsiveness} proposed an adaptive method to adjust a planning horizon time such that prediction of a future state that the robot will likely move toward and planning to said future state can be interleaved in a stable manner.  Also, work by Murray et al. \cite{murray2016robot} accelerated road-map based path planning by creating custom hardware computer chips that check collision states in parallel.  

% This acceleration better accommodates iterative re-planning in time-sensitive, real-time scenarios.    

As mentioned above, motion planning algorithms are effective at getting from a start point to a goal point, but it is difficult for these approaches to enforce what the path does between these boundary points.  In contrast, our method tries to achieve any other motion objectives as best as possible on-the-fly in addition to environment collision avoidance, such as end-effector pose matching over time.      
% \cite{Holmes-RSS-20}

\textit{Trajectory Optimization}---
Trajectory optimization is an approach used to optimize motions to match desired motion qualities (see Betts \cite{betts1998survey} for a review).  Trajectory optimization methods for robot motion, such as CHOMP \cite{ratliff2009chomp}, STOMP \cite{kalakrishnan2011stomp}, and Trajopt \cite{schulman2013finding} include environment collision avoidance techniques; however, the quality and convergence of the computed motion paths using these methods greatly depends on the quality of the initial condition.  Further, these methods often formulate their environment collision-avoidance objective with respect to a pre-computed signed distance field of the environment, which is infeasible to routinely re-compute and update on-the-fly.  Thus, these methods are generally not well suited for real-time, dynamic environments.  In contrast, our method not only accommodates arbitrary motion objectives and constraints, but also accommodate real-time, dynamic environment collision avoidance, at the expense of global optimality.     

%  In this framework, specified \textit{constraints} define the requirements of the motion and specified \textit{objectives} define desired properties of the movement. 

\textit{Inverse kinematics}---
The process of calculating joint values on articulated chains that produce desired pose goals of end-effectors, called inverse kinematics (IK), has been extensively studied in robotics and animation (see Aristidou \cite{aristidou2018inverse} or Nakamura \cite{nakamura1990advanced} for a full review).  A main objective of IK solvers is to reliably match a given end-effector pose goal as quickly
as possible.  A state-of-the-art solver to achieve this central goal on is the optimization-based Trac-IK solver proposed by Beeson and Ames \cite{beeson2015trac}.  Recent work has showed the benefit of incorporating motion \textit{feasibility} objectives, such as self-collision avoidance and singularity avoidance, in addition to the standard motion accuracy objectives mentioned above \cite{rakita2018rss}.  This was shown to be particularly effective for practical use cases such as teleoperation or shared control \cite{rakita2017motion, rakita2019shared, rakita2018autonomous}.  Our current work builds on this concept of providing both motion feasibility and accuracy in real-time inverse kinematics, though we are attempting to extend the definition of motion feasibility to include the avoidance of environment collisions in addition to kinematic singularities, self-collisions, and joint-space discontinuities.     
\section{Technical Overview}
\label{sec:technical_overview}
In this section, we provide background and notation for our problem, and overview our overall method.

\subsection{Notation and Problem Statement}
Suppose $\Theta \in \mathbb{R}^n$ is an $n$-dimensional robot configuration.  Next, consider $\mathbf{f}(\textit{t}, \Theta)$ to be an \textit{objective function} that maps a time value $\textit{t}$ and a robot configuration to some scalar objective output value, $\textit{f} \in \mathbb{R}$.  Note that this implies that the robot's objective can change with respect to time, $\textit{t}$.  The output of the objective function, $\textit{f}$, signifies how well the joint configuration $\Theta$ reflects the objective specified by $\mathbf{f}$ at time $\textit{t}$.  Lastly, consider a set $E$ consisting of obstacles in the environment.  Each of the $K$ obstacles in $E$ will be considered a function $\mathbf{e}_k(\textit{t})$ that maps time $\textit{t}$ to spatial information about the obstacle.  This spatial information could take many forms, such as a signed distance field, occupancy map, triangulated mesh, etc.  Regardless of spatial representation choice, we assume that there is some defined notion of distance between a robot configuration and an obstacle at time $\textit{t}$, which we will denote as $d(\Theta, \mathbf{e}_k(\textit{t}))$.  We intentionally leave these definitions as quite general in this section, and we detail the exact environment spatial representation and definition of distance we used in our method in \S\ref{sec:env_representation}.       

% Suppose $\Theta(\textit{t})$ is a function that maps a time scalar value, $\textit{t}$, to an $n$-dimensional robot configuration, $\theta \in \mathbb{R}^n$.  Next, consider $\mathbf{f}(\textit{t}, \Theta(\textit{t}))$ to be an \textit{objective function} that maps a time value $\textit{t}$ and a robot configuration at time $\textit{t}$ to some scalar objective output value, $\textit{f} \in \mathbb{R}$.  Note that this implies that the robot's objective can change with respect to time, $\textit{t}$.  The output of the objective function, $\textit{f}$, signifies how well the joint configuration $\Theta(\textit{t})$ reflects the objective specified by $\mathbf{f}$ at time $\textit{t}$.  Lastly, consider a set $E$ consisting of obstacles in the environment.  Each of the $K$ obstacles in $E$ will be considered a function $\mathbf{e}_k(\textit{t})$ that maps time $\textit{t}$ to spatial information about the obstacle.  This spatial information could take many forms, such as a signed distance field, occupancy map, triangulated mesh, etc.  Regardless of spatial representation choice, we assume that there is some defined notion of distance between a robot configuration and an obstacle at time $\textit{t}$, which we will denote as $d(\Theta(\textit{t}), \mathbf{e}_k(\textit{t}))$.  We intentionally leave these definitions as quite general in this section, and we detail the exact environment spatial representation and definition of distance we used in our method in \S\ref{}.      

Using the notation above, the problem investigated in this work is to compute a joint configuration $\Theta$ at a given time $\textit{t}$ that minimizes $\mathbf{f}(\textit{t}, \Theta)$, that may include sub-objectives like joint smoothness or matching end-effector pose goals, while maintaining a distance $d(\Theta, \mathbf{e}_k(\textit{t})) > \epsilon, \ \forall k$, where $\epsilon$ is some reasonable cut-off distance between collision and non-collision.  Additionally, the objective $\mathbf{f}$ at time $\textit{t} + \delta$ and the spatial information about each obstacle at time $\textit{t} + \delta$, for any $\delta > 0$, are unknown at time $\textit{t}$.  We also want to solve the above problem as fast as a possible since a fast run-time is essential for many applications.  

% The problem statement above accounts for a single pose optimization for a single objective.  In practice, this problem will typically be solved over a sequence of objectives at monotonically increasing time points to create a sequence of output joint states.  In the case that time progresses ``as fast as possible'', \textit{e.g.,} in a real-time, dynamic setting, our method will begin optimizing a solution at $\textit{t} + \Delta \textit{t}$ as soon as a solution at time $\textit{t}$ was calculated, where $\Delta \textit{t}$ is some processing time it took to compute the solution.  Thus, a secondary challenge that our method must address is keeping this $\Delta \textit{t}$ processing time as small as possible to meet the real-time demands of dynamic environments and objectives. 

% While $\Theta$, $\mathbf{f}$, and $\mathbf{e}$ theoretically represent continuous functions, in practice, we assume that this motion generation process progresses in a discrete manner ``as fast as possible'' at given time points.  For instance, the method begins computing $\Theta(\textit{t} + \Delta \textit{t})$ as soon as $\Theta(\textit{t})$ is calculated, where $\Delta \textit{t}$ is some processing time it took to compute $\Theta(\textit{t})$.  Thus, a secondary challenge that our method must address is keeping this $\Delta \textit{t}$ processing time as small as possible to meet the real-time demands of dynamic environments and objectives.  

\subsection{Non-linear Optimization Structure}
We cast the minimization problem posed above as a constrained non-linear optimization problem:

\begin{equation} \label{eq:opFramework}
\begin{gathered}
\Theta = \argmin_{\Theta} \ \ \textbf{f}(\textit{t}, \Theta) \\
\text{s.t.} \  \textbf{c}(\Theta) \  \geq \  \textbf{0}, \ \ 
l_{i} \leq \Theta_{i} \leq u_{i}, \forall i \quad \quad \quad
\end{gathered}
\end{equation}

Here, $\textbf{c}$ is a set of inequality constraints and $l_{i}$ and $u_{i}$ values define the upper and lower bounds for the robot's joints.     

We express our objective function as a weighted sum of individual motion goals as follows:

\begin{equation}
\begin{gathered}
\label{eq:objective_function}
\textbf{f}(\textit{t}, \Theta) = \sum_{j=1}^{J} \ w_j * f_j(\textit{t}, \Theta, \Omega_j)
%  h_i(\Theta, v(t)) \
\end{gathered}
\end{equation}

Here, $w_j$ is a weight value for each term which sets an importance for a given objective term, and $f_j$ is an objective term function that encodes a single sub-goal, with $\Omega_j$ being model parameters used to construct some loss function.  

To facilitate combining of potentially many objectives, it is important to normalize each term such that their outputs are over a uniform range.  For instance, a term with weight $2$ will ideally hold twice as much importance in the optimization as a term with weight $1$, regardless of the common output ranges of the two terms.  To accomplish this normalization, we use the Groove parametric loss function proposed in prior work \cite{rakita2018rss, rakita2020analysis}, though any loss function that achieves effective normalization of multiple objectives should suffice.  The Groove loss function places a narrow ``groove'' around the goal values, a more gradual falloff away from the groove in order to better integrate with other objectives, and exhibits a consistent gradient that points towards an optimal point. This normalization loss is a Gaussian surrounded by a more gradual polynomial:

\begin{equation}
\label{eq:loss_function}
\begin{gathered}
f_j(\textit{t}, \Theta, \Omega_j) = \\ (-1)^\textit{n} \textit{exp}(\frac{-(\chi_i(\textit{t}, \Theta) - \textit{s})^2}{2\textit{c}^2})
\ + \textit{r}*(\chi_i(\textit{t}, \Theta) - \textit{s})^4
\end{gathered}
\end{equation}

Here, the scalar values $\textit{n}, \textit{s}, \textit{c}, \textit{r}$ form the set of model parameters $\Omega$.  Together, they shape the loss function to express the needs of a certain term.  Here, $\textit{n} \in \{0,1\}$, which dictates whether the Gaussian is positive or negative.  The value $\textit{s}$ shifts the function horizontally, and $\textit{c}$ adjusts the spread of the Gaussian region.  The $\textit{r}$ value adjusts the transition between the polynomial and Gaussian regions, higher values showing a steeper funneling into the Gaussian region and lower values flattening out the boundaries beyond the Gaussian.  The scalar function $\chi$ assigns a numerical value to the current robot configuration that will serve as input to the loss function.

Our method includes seven objective terms and one constraint in the non-linear optimization structure above by default.  The default objective terms encode: (1) end-effector position goal matching; (2) end-effector orientation goal matching; (3) minimized joint velocity; (4) minimized joint acceleration; (5) minimized joint jerk; (6) self-collision avoidance; and (7) environment collision avoidance.  The one constraint is designed to avoid kinematic singularities.  We model objectives 1--6 and the constraint based on prior work \cite{rakita2018rss}.  Derivative information about velocities, accelerations, and jerks are approximated using finite differencing over a short history of prior poses.  We provide details on how we structure and pass spatial information into the environment collision avoidance objective in the following section.  We note that this overall structure is modular and any of the above objectives or constraints can be removed and any additional objectives or constraints can be accommodated.

\section{Technical Details}
\label{sec:technical_details}

In this section, we provide details on how we represent spatial information in our method and how we structure our collision avoidance objective that uses this information. 

\begin{figure}[t!]
	\includegraphics[width=\columnwidth]{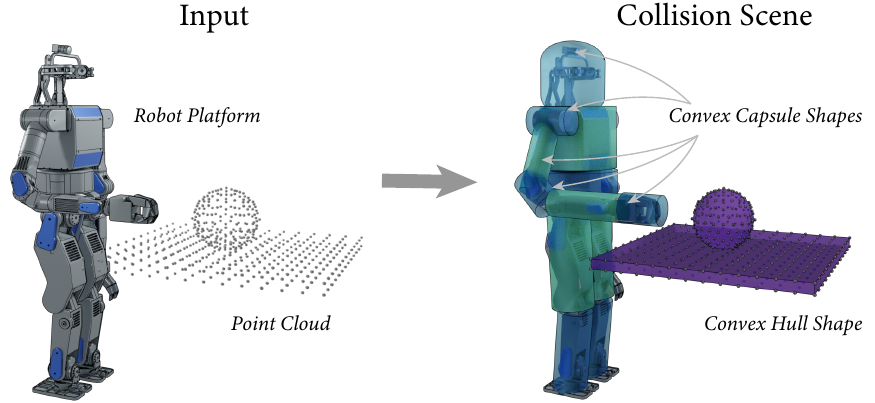}
	\caption{(Left) Our method takes as input a robot platform and set of any collision objects as point clouds.  (Right) Our method converts these inputs into a Collision Scene, comprised of convex capsule shapes around the robot links and convex hull shapes enveloping the point cloud objects.}
	\label{fig:collision_scene}
	\vspace{0pt}
\end{figure}

\subsection{Environment Representation and Structure}
\label{sec:env_representation}
As mentioned above, each collision function, $\mathbf{e}_k(\textit{t})$, maps time to some spatial information about the collision object at that time.  At a high level, our goal is to use convex shape representations that are fast and scalable for computing distances between collision objects and the robot's links.  This technique that has been shown to be effective in many robotics and graphics applications \cite{pan2012fcl, schulman2013finding, ehmann2001accurate}, and we extend these approaches to integrate them within a per-instance optimization.  Our method takes four steps to achieve this goal: (1) Each collision object is input as a point cloud representation; (2) Each point cloud is converted into a convex hull object using the QuickHull algorithm \cite{barber1996quickhull}.  If a particular collision object is not well represented as a convex hull, a decomposition algorithm could break down the object into convex sub-components \cite{larsson2006dynamic, lien2006approximate}; (3) Each convex hull object is updated at each given time $\textit{t}$ using some rigid transformation.  This means that the $\mathbf{e}_k(\textit{t})$ mappings represent rigidly transformed convex hull shapes at time $\textit{t}$; and (4) Each link of the robot is automatically wrapped in a convex shape (a capsule, by default).  Each of these link objects is also updated at each given time $\textit{t}$ based on the robot's forward kinematics model associated with its joint state at that time.  We refer to a function that maps $\textit{t}$ to the rigidly transformed convex shape wrapped around the $n$-th link as $\mathbf{l}_n(\textit{t}, \Theta)$.

We refer to the collection of all all convex hull collision objects and all robot link convex shapes as the \textit{Collision Scene} (illustrated in Figure \ref{fig:collision_scene}).  In the following section, we overview how we use this representation to compute the robot's distance to a collision state at any given time.

% \subsection{Distance to Collision}
% In order to prevent the robot from colliding with the environment, we need some definition of what it means for the robot to be ``close'' to an obstacle.  Our method defines the distance of some robot joint configuration, $\Theta(\textit{t})$, to an obstacle, $\mathbf{e}_k(\textit{t})$, as the minimum distance between the obstacle and one of the links on the robot, $\mathbf{l}_n(\Theta(\textit{t}))$:

\begin{comment}
\begin{equation}
\label{eq:distance_definition}
\begin{gathered}
d(\Theta(\textit{t}), \mathbf{e}_k(t)) = \min_{n} \ \textit{dis}( \mathbf{e}_k(\textit{t}), \mathbf{l}_n(\Theta(\textit{t})) )
\end{gathered}
\end{equation}
\end{comment}

% If this distance is above some threshold $\epsilon$ for all obstacles in the Collision Scene, this implies that the robot is a safe distance from all obstacles.    

\subsection{Environment Collision Avoidance Objective Term}
\label{sec:objective_term}
In order to discourage the robot from colliding with the environment, it is necessary to define of what it means for the robot to be ``close'' to an obstacle.  We encode a distance to a collision state using the following cost function:

\begin{equation}
\label{eq:cost_function}
\begin{gathered}
\chi_{c}(\textit{t}, \Theta(\textit{t})) = \sum_{\mathbf{e_k \in \mathcal{A}}} \sum_{n}^N \frac{(5\epsilon)^2}{\textit{dis}( \mathbf{e}_k(\textit{t}), \mathbf{l}_n(\textit{t}, \Theta) )^2}
\end{gathered}
\end{equation}

Here, $\epsilon$ is some scalar value that signifies a reasonable cutoff distance between collision and non-collision.  For example, in our prototype system, we use a value of $\epsilon = 0.02$ (represented in meters).  The $\textit{dis}$ function computes the shortest straight-line distance between the input shapes, \textit{i.e.}, a collision object and one of the robot's links.  We compute the $\textit{dis}$ function using a Support Mapping computation, as this is an efficient way to find the shortest distance between two convex shapes \cite{kenwright2015generic}.  Lastly, $\mathcal{A}$ is a set of ``active'' collision objects at the given time, $\textit{t}$.  Calculating this sum over just a subset of all collision objects maintains the efficiency and scalability of this computation when many obstacles are present in the Collision Scene.  We overview how we filter the Collision Scene to select a subset of salient collision objects in the following section.

The cost function in Equation \ref{eq:cost_function} was designed to be smooth and differentiable, as opposed to, for example, taking the minimum distance, such that it effectively mixes into a multi-objective, gradient-based optimization structure.  We incorporate the cost function in Equation \ref{eq:cost_function} into our optimization framework as a single objective term using the loss function in Equation \ref{eq:loss_function}.  In our prototytpe system, we used Groove loss parameters of $\textit{n} = 1$, $\textit{s} = 0$, $\textit{c} = 2.5$, and $\textit{r} = 0.0035$. These values were selected to reflect the standard output range of the cost function.  In particular, these loss function parameters ensure that this term's objective output significantly ramps up when the robot approaches a distance $\epsilon$ from an obstacle.  Note that the cost function output is high when the robot is close to collision, thus the goal Gaussian region is negative in the loss function.    

% While the definition in Equation \ref{eq:distance_definition} is precise, it is not amenable for a gradient-based optimization loop since it is non-differentiable.  In this section, we show how we convert this distance to a differentiable cost function that achieves the same effect and demonstrate how it is incorporated into the optimization structure specified above.     

% We compute the $\textit{dis}$ function using a Support Mapping representation, as this is an efficient way to find the shortest distance between two convex shapes \cite{kenwright2015generic}.  

\subsection{Collision Scene Filtering}
\label{sec:collision_scene_filtering}
As mentioned above, we only compute Equation \ref{eq:cost_function} over a set of ``active'' obstacles in set $\mathcal{A}$.  Filtering the obstacles and only considering a subset at any given time prevents the collision avoidance objective term from becoming too slow, especially in the presence of many environment obstacles.  At every given time $\textit{t}$, our method starts with all collision objects set as active and prunes this set based on two criteria: (1) All obstacles with a distance greater than some margin, $\Upsilon$, from all of the robot's links at a given time will be removed from the active set; and (2) If $| \mathcal{A}| > N$ after step 1, only the $N$ obstacles that have the highest cost according to Equation \ref{eq:cost_function} are active.  All other obstacles are removed from $\mathcal{A}$.            

Our method efficiently achieves criterion 1 above by using a broad-phase and narrow-phase collision detection pipeline.  The broad-phase step uses axis-aligned bounding box (AABB) hierarchies to quickly disregard obstacles that are guaranteed to not be within a distance of $\Upsilon$.  This phase uses a Dynamic Bounding Volume Tree to efficiently store, update, and query collision information, even in a dynamic environment.  The narrow-phase performs ground truth distance checking only on the obstacles that were not culled by the broad-phase.  Our prototype system uses an $\Upsilon$ margin distance of one meter, and sets the maximum number of active obstacles, $N$, as three.

\section{Evaluation}
\label{sec:evaluation}

In this section, we overview our experiments designed to assess the efficacy of our method.      

\begin{figure*}[t!]
	\includegraphics[width=\textwidth]{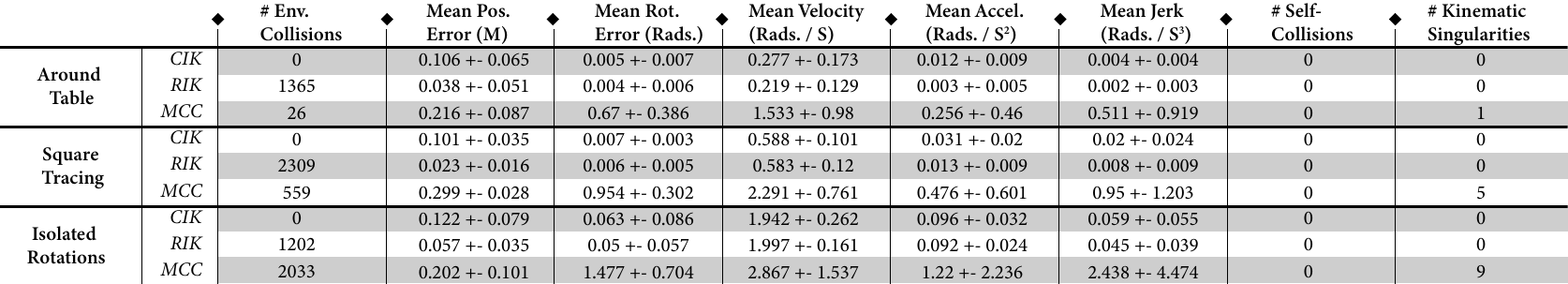}
	\caption{Results from Experiment 1.  Range values represent standard deviation.}
	\label{fig:results_experiment1}
	\vspace{-10pt}
\end{figure*}

\subsection{Implementation Details}
\label{sec:implementation_details}
A prototype implementation of our method was configured as an extension of the Rust version of the \textit{RelaxedIK} library\footnote{\href{https://github.com/uwgraphics/relaxed_ik}{https://github.com/uwgraphics/relaxed\_ik}}.  Spatial data structures relating to the broad and narrow phase collision checking and convex hull conversions all use the \texttt{ncollide3d} library\footnote{\href{https://ncollide.org/}{https://ncollide.org/}}.  Our method uses the proximal averaged Newton-type Method (PANOC) optimization approach \cite{stella2017simple}, implemented in the Rust Optimization Engine (OpEN) library.  Our approach also works well with a variety of non-linear solvers, such as those offered by NLopt.  All gradients in our work were computed using finite differencing.  All evaluations were run on an AMD Ryzen 7 2700X Processor (3.70GHz) with 16GB RAM.    

% All evaluations were run on a Lenovo Legion laptop with an i7-9750H processor and 32GB RAM.

% \footnote{\href{https://alphaville.github.io/optimization-engine/}{https://alphaville.github.io/optimization-engine/}}

\subsection{Evaluation Benchmark}
We developed a set of three benchmark tasks to compare our method against alternative approaches.  We will refer to these tasks as \textit{Around Table}, \textit{Square Tracing}, and \textit{Isolated Rotations}. 

The \textit{Around Table} task involves the robot's end-effector being driven toward and through a table surface.  The robot's goal in this task is to avoid colliding with the table and continue following the goal trajectory on the other side.  The \textit{Square Tracing} task involves the robot's end-effector tracing a square shape.  A dynamic cube object approaches the robot on its way around the square, and the robot's goal is to avoid colliding with the cube and continue to follow the perimeter of the square as best as possible.  Lastly, the \textit{Isolated Rotations} task involves the robot's end-effector position remaining static, and the robot rotates its end-effector 90 degrees around all of its primary axes in sequence.  While these rotations are happening, three dynamic sphere objects in the environment continuously encroach on the robot's space, and the goal is to avoid colliding with these spheres while still matching the specified end-effector rotations and static position point as best as possible.        

The tasks listed above were all run on five simulated robots: a Universal Robots UR5 (6-DOF), a Rethink Robotics Sawyer (7-DOF), a Kinova Jaco (7-DOF), a Kuka IIWA (7-DOF), and the torso and right arm of the Rainbow Robotics Hubo+  (8-DOF).  Each task was run 10 times per robot for every condition in our benchmark. 

% \footnote{\href{http://www.rethinkrobotics.com/sawyer/}{http://www.rethinkrobotics.com/sawyer/}}
% \footnote{\href{https://www.kinovarobotics.com/en}{https://www.kinovarobotics.com/en}}
% \footnote{\href{https://www.kuka.com/en-us/products/robotics-systems/industrial-robots/lbr-iiwa}{https://www.kuka.com/en-us/products/robotics-systems/lbr-iiwa}}
% \footnote{\href{http://www.rainbow-robotics.com/new/}{http://www.rainbow-robotics.com/new/}}
% \footnote{\href{https://www.universal-robots.com/products/ur5-robot/}{https://www.universal-robots.com/products/ur5-robot/}}

\subsection{Evaluation Measures}
We assessed eight primary measures in our evaluations: mean position error (meters), mean rotational error (radians), mean joint velocity ($rad/s$), mean joint acceleration ($rad/s^2$), mean joint jerk ($rad/s^3$), total number of kinematic singularities, total number of self-collisions, and total number of environment collisions.

% We also report on the average solution time (seconds) for each condition.  

% Our benchmark consists of two tasks on a Universal Robots UR5 (6-DOF), two tasks on a Rethink Robotics Sawyer (7-DOF), and three bimanual tasks on the DRC-Hubo+ (15-DOF).

\subsection{Experiment 1: Comparisons with Alternative Approaches}
In our first experiment, we compared our method to two alternative approaches: (1) \textit{RelaxedIK} \cite{rakita2018rss} (which we will refer to as \textit{RIK}); and (2) the MoveIt Cartesian Controller \cite{chitta2012moveit} (which we will refer to as \textit{MCC}).  The \textit{RIK} condition uses the open-source Rust version of RelaxedIK.  It includes many objective terms in its objective function, such as end-effector pose matching, smooth joint motion, and self-collision avoidance, but does not include the environment collision avoidance methods discussed in this work.  

\begin{figure*}[t!]
	\includegraphics[width=\textwidth]{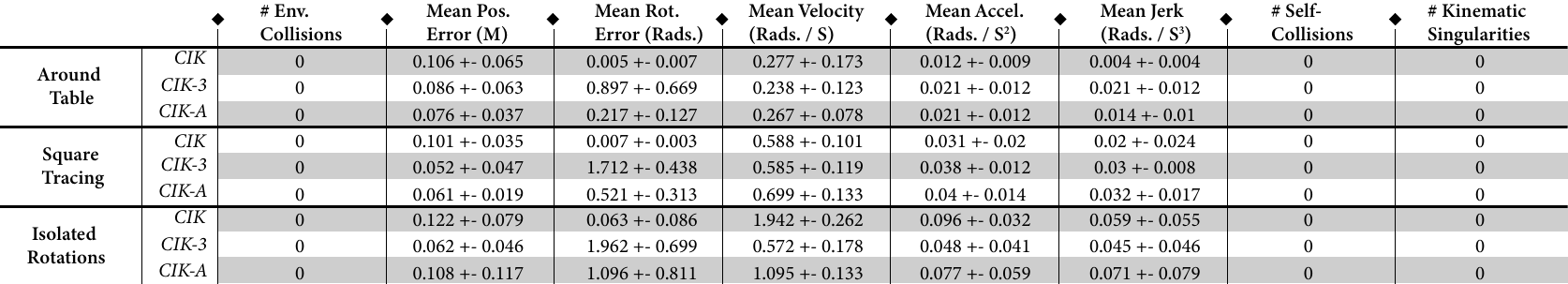}
	\caption{Results from Experiment 3.  Range values represent standard deviation.}
	\label{fig:results_experiment3}
	\vspace{-10pt}
\end{figure*}

The \textit{MCC} controller uses the open-source MoveIt library.  It involves three steps: (1) Compute a configuration that matches a desired end-effector pose goal using Trac-IK \cite{beeson2015trac}; (2) Compute a feasible, collision-free path to the configuration found in step 1 using the RRT-Connect planner \cite{kuffner2000rrt}; and (3) Move along the path found in step 2, and return to step 1 once the robot either completes the path or hits a dynamic collision along the path.  If the planner ever does not find a feasible path to the given configuration in step 2, the robot maintains its current configuration, time advances some fixed step, and the approach goes back to step 1. In our evaluation, we ``pause'' the time parameter $\textit{t}$ on the trajectory and obstacles during step 2.  Thus, this allows the planner to be ``infinitely fast'' from the perspective of the input trajectory and environment.  We made this evaluative decision because our goal was to assess the general \textit{MCC} approach under ideal conditions rather than potentially show that one implementation of one particular planner is too slow to keep up with the real-time demands of our benchmark.    

Our results from Experiment 1 are summarized in Figure \ref{fig:results_experiment1}.  We see that our method avoids collisions with the environment more effectively than both \textit{RIK} and \textit{MCC}.  Also, our method achieves smoother motion and reflects the given end-effector pose goals more precisely than \textit{MCC}.  While \textit{RIK} does avoid self-collisions and kinematic singularities (as reported in prior work), and reflects lower end-effector pose errors than our method, this extra accuracy comes at the high cost of many collisions with the environment.   

\subsection{Experiment 2: Local Minimum Test}
In our second experiment, our goal was to assess if our method is prone to getting stuck in local minimum regions.  This assessment involved moving the table closer to the robot in the \textit{Around Table} task, such that the end-effector was guided more through the center of the table rather than close to the edge.  We compared our method to the \textit{MCC} approach described above on this task.

\begin{figure}[t!]
	\includegraphics[width=\columnwidth]{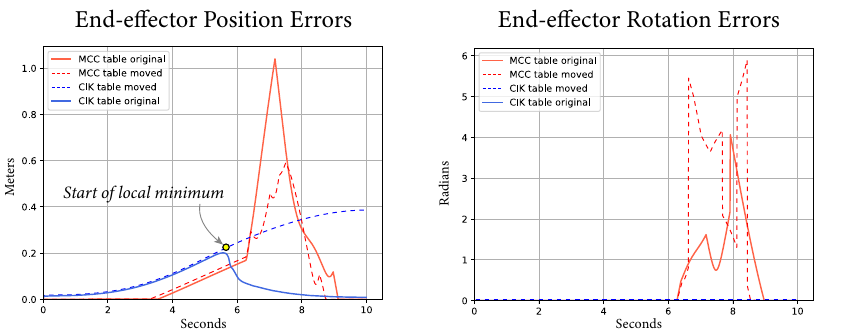}
	\caption{End-effector position and rotation errors from our local minimum experiment.  (Left) We see that our method falls into a local minimum when the table is moved closer to the robot.  While \textit{MCC} gets around the table using its global planning capibilities, this comes at the cost of high end-effector position and rotation errors.  }
	\label{fig:local_minimum}
	\vspace{0pt}
\end{figure}

While our method still avoids colliding with the table, it does not ultimately find a way to get around the table to the other side.  Thus, we see that our method is prone to falling into local minima, while other global planners do not get stuck in such regions.  We note that while \textit{MCC} did find a way around the table, this came at the expense of considerably high end-effector position and rotation errors.  A graph of end-effector translation errors of the two conditions can be seen in Figure \ref{fig:local_minimum}.  

\subsection{Experiment 3: Assessment of Objective Term Variants}
In our third experiment, we compare results on our evaluation benchmark with adjusted objectives.  This experiment was designed to show the flexibility of our approach in terms of accommodating different or dynamically changing objectives.  Our first condition, \textit{CIK}, uses the same position and orientation matching objectives as Experiments 1 and 2 above.  Our second condition, \textit{CIK-3}, only considers end-effector position goal matching, and does not try to match the orientation of goals as well.  Lastly, our third condition, \textit{CIK-A}, adaptively adjusts the weight on the orientation matching objectives on-the-fly, such that the importance of orientation matching is reduced when the robot is close to a collision state and raised to its standard value when the robot is not close to a collision.  

Our results from Experiment 3 can be seen in Figure \ref{fig:results_experiment3}.  We see that all of the evaluated variants of our method achieve their respective objectives, all while avoiding collisions with the environment.  For example \textit{CIK-3} reliably matches end-effector position goals, though it has a very high orientation matching error, while \textit{CIK-A} achieves some orientation matching accuracy while still achieving higher positional accuracy than the baseline \textit{CIK}.  This demonstrates that our method is able to reflect a variety of different objectives, even those that dynamically change in real-time.      

\begin{comment}
\begin{table}[t!]
	\includegraphics[width=\columnwidth]{figures/figures-01.png}
	\caption{}
	\label{tab:performance}
	\vspace{3pt}
\end{table}
\end{comment}

\subsection{Experiment 4: Performance and Scalability Testing}
In our final experiment, we tested the performance and scalability of our method in a real-time control setting.  This experiment involved a user interactively driving the end-effector position and orientation of a simulated Sawyer robot in a ROS RViz environment using a keyboard controller.  Obstacles placed in the environment were also able to be interactively moved by the user.  Our ability to test our method on a real robot using a more robust control interface was hindered by the SARS-CoV2 pandemic.  

Our first interactive test environment involved four sphere obstacles placed around a simulated robot.  Our method avoided collisions with all obstacles during this test, and average run-time performance for this environment was approximately 500 microseconds per solve (2,000 Hz).  Our second test environment was designed to test the scalability of our method, and involved 100 Stanford bunnies, each with over 30,000 vertices, around the environment.  Each Stanford bunny took 18 milliseconds on average to compute a convex hull using the QuickHull algorithm.  Our method again avoided collisions with all obstacles during this test, and average run-time performance for this environment was approximately 16 milliseconds per solve (60 Hz).
\section{Discussion}
\label{sec:discussion}
In this work, we presented a method for generating robot motions on-the-fly that avoid collisions with static or dynamic obstacles in an environment, while also simultaneously accommodating any other motion objectives.  In this section, we discuss limitations and implications of our work.    

% Through three empirical experiments, we demonstrated the efficacy of our work compared to other approaches.  
% We overviewed how to effectively include environment collision avoidance as an objective in a multi-objective non-linear optimization structure, and we provide first solutions for how to effectively pass spatial information from the scene to this performance critical optimization.

\subsection{Limitations}
We note a number of limitations of our work that suggest future extensions.  First, our method is prone to becoming trapped in local minima regions, as seen in Experiment 2.  Further investigation is required to characterize when our method is susceptible to falling into local minima, and extensions of our work could explore ways to infuse more global path planning techniques within our optimization-based structure to get out of these minima \cite{lavalle2006planning}.  Also, our work was tested in simulation with synthetic data and does not consider the challenges of real-world sensing or noisy data.  Extensions of our work could explore how to efficiently accommodate sensed data form the environment and how to incorporate confidence bounds into our objectives in the case of noisy sensors.          

% Second, while our current spatial representation above achieves efficiency, in part, by wrapping individual point cloud objects into convex hulls, this implies that this technique will not currently work on non-convex objects.  Extensions of our work could use decomposition methods, such as Hierarchical Approximate Convex Decomposition (HACD), to break down collision objects into convex sub-components.  

% Extensions of our work could investigate efficient ways to store and access points in non-convex objects or how to automatically split non-convex shapes into convex segments to be compatible with our current techniques.  

\subsection{Implications}
We believe that our method could benefit various areas of robotics, such as teleoperation, shared-control, and reinforcement learning.  For example, teleoperation and shared-control interfaces could utilize our method in home health-care, telenursing, or nuclear materials handling applications to mitigate collisions while still achieving other goals as best as possible.  Also, unsupervised or semi-supervised learning agents could explore and manipulate their environments without high risk of collisions when forming their policies.  We plan to explore these directions in future work. 

% Also, unsupervised or semi-supervised learning agents could explore and manipulate their environments without fear of collisions in order to build up more robust policies.  

% \section*{ACKNOWLEDGMENT}
% Omitted for review process.
% This research was supported by a Microsoft Research PhD Fellowship, National Science Foundation award 1830242, and NASA Cooperative Agreement 80NSSC19M0124.

%%%%%%%%%%%%%%%%%%%%%%%%%%%%%%%%%%%%%%%%%%%%%%%%%%%%%%%%%%%%%%%%%%%%%%%%%%%%%%%%

\balance
\bibliographystyle{IEEEtran}
\bibliography{references}

\end{document}